\documentclass[11pt]{article}
\usepackage[preprint]{acl}        

\usepackage[T1]{fontenc}
\usepackage[utf8]{inputenc}
\usepackage{microtype}
\usepackage{tipa}             
\usepackage{newunicodechar}

\DeclareUnicodeCharacter{1EB9}{\textsubdot{e}} % ẹ
\DeclareUnicodeCharacter{1EB8}{\textsubdot{E}} % Ẹ
\DeclareUnicodeCharacter{1ECD}{\textsubdot{o}} % ọ
\DeclareUnicodeCharacter{1ECC}{\textsubdot{O}} % Ọ
\DeclareUnicodeCharacter{1E63}{\textsubdot{s}} % ṣ
\DeclareUnicodeCharacter{1E62}{\textsubdot{S}} % Ṣ

\DeclareUnicodeCharacter{FFFD}{}

\usepackage{booktabs}
\usepackage{graphicx}
\usepackage{tabularx}
\usepackage{array}
\usepackage{multirow}
\usepackage[table]{xcolor}
\usepackage{amssymb}
\usepackage{xspace}
\usepackage{algorithm}
\usepackage{algpseudocode}
\usepackage{subfigure}          
\usepackage{lineno}
\usepackage{url}
\usepackage{hyperref}            
\usepackage[normalem]{ulem}

\usepackage{worldflags}
\newcommand{\smallflag}[1]{\worldflag[width=0.3cm]{#1}}

\newcommand{\afri}{\textsc{AfriCaption}\xspace}

\title{\afri: Establishing a New Paradigm for Image Captioning in African Languages}

\author{
  Mardiyyah Oduwole\textsuperscript{1}\thanks{Equal contribution.} \quad
  Prince Mireku\textsuperscript{1,2}\footnotemark[1] \quad
  Fatimo Adebanjo\textsuperscript{1}\thanks{Equal contribution.} \quad
  Oluwatosin Olajide\textsuperscript{1}\footnotemark[2] \\
  \textbf{Mahi Aminu Aliyu}\textsuperscript{1,3} \quad
  \textbf{Jekaterina Novikova}\textsuperscript{1}\thanks{Supervisor.} \\[1em]
  \small
  \textsuperscript{1}ML Collective \quad
  \textsuperscript{2}Ashesi University \quad
  \textsuperscript{3}Abubakar Tafawa Balewa University \\
  \texttt{mardiyyah.oduwole@mlcollective.org}
}

\begin{document}
\maketitle

% \begin{document}
\maketitle

\maketitle

\begin{abstract}
Multimodal AI research has overwhelmingly focused on high-resource languages, hindering the democratization of advancements in the field. To address this, we present AfriCaption, a comprehensive framework for multilingual image captioning in 20 African languages and our contributions are threefold: (i) a curated dataset built on Flickr8k, featuring semantically aligned captions generated via a context-aware selection and translation process; (ii) a dynamic, context-preserving pipeline that ensures ongoing quality through model ensembling and adaptive substitution; and (iii) the AfriCaption model, a 0.5B parameter vision-to-text architecture that integrates SigLIP and NLLB200 for caption generation across under-represented languages. This unified framework ensures ongoing data quality and establishes the first scalable image-captioning resource for under-represented African languages, laying the groundwork for truly inclusive multimodal AI.
\end{abstract}

\section{Introduction}
The digital divide in multimodal AI is starkly evident, with most advancements centered around a selected few Western languages, leaving non-Western languages, especially African languages, underrepresented ~\citep{longpre2024bridging}. This under-representation perpetuates a cycle of exclusion, where machine learning systems fail to generalize to global contexts and perform poorly for speakers of low-resource languages, creating a barrier to inclusive AI development.

Datasets and models have both been two strong pillars of machine learning since its inception, where the performance of a good model stems not only from its architecture or training hyperparameters but also from the foundational dataset on which it is trained. %However, the majority of existing datasets, such as MS COCO \citep{DBLP:journals/corr/LinMBHPRDZ14}, Flickr8k \citep{Hodosh2013Flickr8K}, and Visual Genome \citep{DBLP:journals/corr/KrishnaZGJHKCKL16}, are predominantly in English, with only a limited selection of translations for other, typically high-resourced, languages. This lack of linguistic diversity in datasets results in the lack of machine learning models available for underrepresented languages, and it reinforces biases and excludes entire populations from benefiting from advances in AI technologies.
% Multimodal datasets and pre-trained models have been instrumental in advancing AI systems capable of understanding both visual and textual information. 

Early benchmark datasets such as ImageNet \citep{5206848} revolutionized computer vision by providing large-scale annotated images, enabling the development of deep learning models. Similarly, MS COCO \citep{DBLP:journals/corr/LinMBHPRDZ14}, Flickr8k \citep{Hodosh2013Flickr8K}, and Visual Genome \citep{DBLP:journals/corr/KrishnaZGJHKCKL16} provided diverse image-text pairings, facilitating advancements in vision-language tasks like image captioning and visual question-answering. However, these datasets are overwhelmingly monolingual, primarily in English, reflecting an inherent bias in AI research \citep{DBLP:journals/corr/abs-1803-09010}. The consequence of this linguistic homogeneity is a failure to generalize AI models across non-Western contexts, limiting their usability and fairness \citep{10.1145/3442188.3445922}.

\begin{figure*}[t]
  \centering
  \includegraphics[width=1\textwidth]{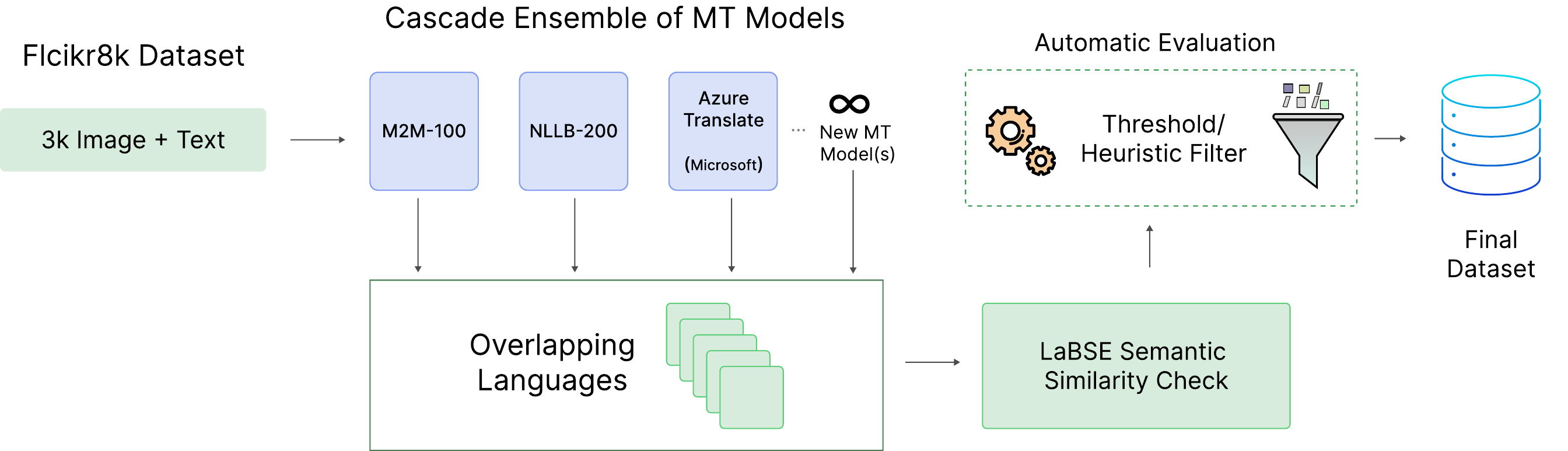}
  \caption{The context-preserving adaptive pipeline, ensuring continuous improvement and high data quality.}
  \label{fig:pipeline}
\end{figure*}

To this end, we introduce \afri: an image captioning model and dataset for African languages. \afri provides an image-text pair dataset and an image captioning model that, in addition to English, covers 20 African languages, spanning across several language families and regions. To the best of our knowledge, this is the first image captioning model and curated caption corpus of this scale built for African languages.  
The key contributions of our work include:

\begin{itemize}
    \item \textbf{The \afri dataset containing diversified multilingual captions}: We create a corpus of human-readable captions in linguistically diverse African languages, including Igbo,  Hausa, Ewe, Yoruba, Luganda, Kinyarwanda, and others spanning Afro-Asiatic, Niger-Congo, and Nilo-Saharan families. \afri addresses the lack of coverage for low-resource African languages, creating opportunities to train and evaluate models while ensuring  linguistic AI representation (Section~\ref{sec:dataset}).

    \item \textbf{Context-preserving pipeline:} We develop and present a novel caption translation process to ensure the African language captions remain faithful to the image semantics (Figure~\ref{fig:pipeline} and Section~\ref{sec:pipeline}).

    \item \textbf{The \afri image captioning model}: We introduce the first image captioning model designed to generate captions in a wide range of African languages. It is the first to support image captioning for the majority of the 20 African languages covered in our dataset. The model aligns a vision encoder (SIGLIP) with a multilingual text decoder (NLLB) to produce captions across these languages (Section~\ref{sec:model}).
\end{itemize}

With this work, we hope to broaden the scope of research and democratise AI, ensuring that cutting-edge technologies benefit a global community rather than just speakers of high-resource languages.

\section{Related Work}

Recent efforts have sought to address the under-representation of diverse languages in multimodal AI. A prime example is OpenAI’s CLIP model \citep{DBLP:journals/corr/abs-2103-00020}, which aligns text and images using large-scale datasets primarily in high-resource languages such as english. While CLIP has demonstrated impressive zero-shot learning capabilities, it struggles to generalize across diverse linguistic contexts, particularly for under-represented languages, such as those spoken across Africa. Similarly, multilingual models such as mBERT \citep{devlin-etal-2019-bert}, XLM-R \citep{conneau2020unsupervisedcrosslingualrepresentationlearning}, and M2M-100 \citep{fan2020englishcentric, schwenk2019ccmatrix, el2019massive} have demonstrated the feasibility of training AI systems across multiple languages, but their reliance on textual corpora that often exclude low-resource languages results in suboptimal performance for African languages  \citep{kakwani-etal-2020-indicnlpsuite}.

One of the more recent efforts to bridge this gap is AViLaMa \citep{sartifyllc2023africanvision}, a large open-source text-vision alignment pre-training model specifically targeting African languages. AViLaMa integrates supervision from several African languages, including Swahili, Hausa, Yoruba, Igbo, Zulu, Shona, Arabic, and Amharic, alongside Western languages, such as English, French, and Portuguese. 
% It expands upon OpenAI’s CLIP by introducing African languages, using techniques like agnostic language encoding and specialized data filtering to improve contextual understanding that standard machine translation often fails to capture.
% The model represents an important step toward addressing the under-representation of African languages in multimodal tasks.

Despite these advancements, a key limitation remains: many of these datasets and models focus heavily on a small subset of widely spoken African languages, often neglecting lower-resource languages that are equally important for the democratization of AI. Additionally, many multimodal models primarily focus on text-vision alignment without addressing the full spectrum of African linguistic diversity in image captioning and other multimodal settings. Although datasets like Multi30k \citep{DBLP:journals/corr/ElliottFSS16} and the CrissCrossed Captions Dataset (CxC) \citep{DBLP:journals/corr/abs-2004-15020} have expanded multilingual representation, they still lack sufficient African language inclusion.

% To advance African language representation in multimodal AI, several initiatives have emerged. Masakhane \citep{inproceedings} has compiled and translated textual datasets for African languages, demonstrating the feasibility of inclusive AI. Similarly, models like AfriBERTa \citep{ogueji-etal-2021-small} and LADL (Low-Resource African Dialect Languages Dataset) \citep{adebara-abdul-mageed-2022-towards} have been developed to support African-language NLP, albeit with limited multimodal integration.

In our work, we aim to contribute to the inclusion of African languages in the advances in the multimodal domain by introducing \afri, the first image captioning model and dataset that broadens the linguistic spectrum and includes 20 African languages. This ensures a wider representation, particularly for under-represented languages. Unlike previous efforts that focus solely on text-vision alignment, our dataset integrates both text and image pairs from the well-known Flickr8k dataset {\citep{Hodosh2013Flickr8K}}, and our model integrates SigLIP's \citep{zhai2023sigmoid} vision encoder with NLLB's decoder \citep{costa2022no}, providing a richer multimodal resource. We also prioritize and linguistic inclusivity by ensuring captions are contextually relevant to each language, fostering nuanced interactions with diverse linguistic groups. We aim to facilitate research in multilingual multimodal AI for low-resource languages and enable models to generalize better across different languages, contributing to a more inclusive global AI landscape.

\section{Background}
\label{sec:background}

Africa is one of the most linguistically diverse regions in the world. Estimates suggest that the continent is home to between 1,500 and 2,000 distinct languages \citep{ethnologue2022}. \footnote{The Ethnologue reports a similar range, though numbers vary with new surveys.} These languages span several major families, including Niger-Congo, Afro-Asiatic, Nilo-Saharan, and Khoisan, and exhibit a wide range of morphological structures. In many African languages, particularly within the Afro-Asiatic and Niger-Congo families, the morphology can be highly complex. 

% Agglutinative and polysynthetic structures are common, as seen in languages such as Amharic, Somali, and several Bantu languages, where affixation and tonal variations create rich and nuanced grammatical systems \citep{longpre2024bridging}.
Many African languages remain absent from the corpora despite being spoken by tens of millions and this has led to the challenge of the significant gap in machine learning (ML) resources. For example, while languages like Yoruba, Hausa, and Igbo, which collectively have speaker populations ranging from approximately 20 to 50 million, are included in some ML datasets, the vast majority of Africa’s languages receive little or no attention \citep{zdnetWorldJust, EqualyzAI_2025}.

\begin{table*}[t]
\centering
\small
\label{tab:language_stats}
\begin{tabularx}{\linewidth}{>{\raggedright\arraybackslash}Xcccc}
\toprule
\textbf{Dataset Name} & \textbf{\#Samples} & \textbf{\#lang} & \textbf{Include African Lang} & \textbf{\#African Langs} \\
\midrule
Multi30K (\citep{DBLP:journals/corr/ElliottFSS16}) & 30,014 & 2 & $\times$ & - \\
Crossmodal-3600 \citep{thapliyal2022crossmodal3600massivelymultilingualmultimodal} & 3,600 & 36 & $\times$ & - \\
COCO-CN \citep{li2019cococn} & 20,342 & 2 & $\times$ & - \\
WIT \citep{srinivasan2021wit} & 11.5M & 108 & \checkmark & unspecified \\
ArtELingo-28 \citep{mohamed2024artelingo28} & 2,000 & 28 & \checkmark & 10+ \\
\midrule
\textbf{\afri (Ours*)} & 8K & 21 & \checkmark & 20 \\
\bottomrule
\end{tabularx}
\caption{Comparison of multilingual image-text datasets with respect to African language coverage. \afri (Ours) is the only dataset to explicitly support 20 African languages, providing broader coverage than existing benchmarks.}
\end{table*}

\subsection{Harms of Misrepresentation}
\label{subsec:harms}
The underrepresentation of African languages in ML datasets and models has significant technical and societal consequences. From a technical perspective, models trained predominantly on high-resource languages fail to generalize to the unique grammatical structures and contexts inherent to many African languages. This can lead to degraded performance, misinterpretation of idiomatic expressions, and ultimately, erroneous outputs when these models are applied in real-world settings. The issue is compounded by the morphological complexity of many African languages, which demands tailored linguistic models that can capture inflectional nuances and tonal variations \citep{kandybowicz2018african}. The harms, however, extend far beyond technical inadequacies. When languages spoken by millions are marginalized in AI research, speakers of these languages are effectively excluded from the benefits of modern technology.

\section{Context-Preserving Pipeline}
\label{sec:pipeline}
A dedicated and reliable system for obtaining quality data in the context of African NLP is a rarely explored topic. This is a result of some MTs performing better in some languages and poorly in others. To this end, we develop a simple and effective pipeline (Figure~\ref{fig:pipeline}) that ensures
data quality and continual updates through a method of model ensembling and substitution.

We start by using the Flickr8k dataset~\citep{Hodosh2013Flickr8K} as input to several publicly available machine translation (MT) models and build a cascade of ensembles where each model is capable of translating at least one African language. We generate and evaluate similarities between embeddings using language-agnostic BERT (LaBSE)\citep{feng2020language}, compared to other methods like back-translation from the target language to English, eliminates the need for computationally expensive resources.

For languages supported by multiple models, we measure the cosine similarity between their translations and retain the version with the highest score in the final dataset. To ensure ongoing quality, we introduce a dynamic replacement mechanism: when newer models outperform previous ones, that is, yielding higher similarity scores, the corresponding translations are updated accordingly. The novelty of our approach lies in the flexibility of our framework, which allows swapping target language translations when better-performing models become available, ensuring continuous improvement without compromising data integrity.

%\textbf{Heuristic Filters}
%\quad
As a final step to ensure the quality of the dataset, we used heuristic filters to filter out suspicious translations. Inspired by the prior work on unified text-to-text transformer \citep{raffel2020exploring} and work on developing heuristic filters through data inspection \citep{penedo2025fineweb}, we devise a mechanism to select close to accurate translations.%Prior work focused on developing heuristic filters through data inspection \citep{penedo2025fineweb}, including manual approaches. In this work, 
We implement a method of manually inspecting and eliminating translations that fall below a set threshold. From observation in translated languages, manual human evaluation in a sample revealed translations that adequately describe an image without loss of context, had cosine similarity scores above 0.53. We then used this as the threshold to select quality translations.

For our dataset, the final collection \( D \) consists of translations \( t \) such that the similarity score \( d(t) \) lies in the interval \([0.53, 0.98]\):
\[
D = \{ t \mid d(t) \in [0.53, 0.98] \}
\] \\
This method allows for data quality assurance even for languages with limited access to human evaluators, which is crucial for creating datasets of under-represented languages with low resources.

%%%%%%%%%%%%%%%%%%%%%%%%%%%%%%%%%%%%%%%%%%%%%%%%
% Dataset: Method, Results, Analysis
%%%%%%%%%%%%%%%%%%%%%%%%%%%%%%%%%%%%%%%%%%%%%%%%
\section{\afri Dataset}
\label{sec:dataset}

% A Sample dataset was shown here

\subsection{Data Selection and Translation}
% Our goal with creating \afri is to experiment with a sufficiently small sample data before scaling the dataset to include large samples. As a result, 
The Flickr8k dataset, which consists of 8,000 images, each accompanied by five human-generated captions, was chosen  particularly for its dense captions upon human review of a few samples.

\textbf{Caption Selection}
\label{sec:caption-selection}
\quad
To compress the dataset to as minimal as possible, a single caption had to be selected among the five to represent a single image. We assume that the best captions will have semantic similarities when compared with each other. We compute a cosine similarity score between all the pairs of captions, and the caption with the highest score is selected in order to avoid potential biases inherent in any singular selection method. In order to ensure preservation of context, which is vital for multilingual tasks, we utilize the pre-trained SentenceBERT model from the sentence transformer \citep{reimers-2019-sentence-bert} to generate vector representations of the captions.

\textbf{Translation Process}
\quad
Leveraging the individual translation capability of multiple machine translators, we experimented with SoTA models that support African languages, proven by literature to have a par performance. Our experiment utilized NLLB200 \citep{nllbteam2022languageleftbehindscaling}, M2M100 \citep{fan2020englishcentric} and Azure Translate \citep{microsoft_azure_translator}, of which the first two are publicly available models. 

\subsection{Quality Assurance}
To assess the quality of translations in \afri, we adopted a two-pronged approach: (1) an automated similarity evaluation using a back-translation method and (2) a human evaluation to ensure contextual fidelity.

\textbf{Automatic Evaluation}
The automatic process leveraged a back-translation strategy using the NLLB200 \citep{nllbteam2022languageleftbehindscaling} MT model. We translated captions from English to target languages and then back to English. The cosine similarity score between the original caption and the back-translated caption was calculated between the embeddings of both the original and translated captions. To preserve context, embeddings were generated using the SentenceBERT model as described in our caption selection process \ref{sec:caption-selection}.

\textbf{Human Evaluation}
To complement automatic evaluation, we conducted a human evaluation study on four languages: Yoruba, Igbo, Hausa, and Ewe. We chose these languages based on proximity to communities where these languages are spoken and to cover a mix of high vs. low-resource scenarios. Yoruba and Hausa are widely spoken and have relatively better MT support (we used Azure and NLLB for Yoruba, NLLB for Hausa), whereas Igbo and Ewe are less supported (both used NLLB; Ewe is especially low-resource). 

We gathered responses from native speakers of these languages. In total, 102 Yoruba, 38 Igbo, 37 Hausa, and 2 Ewe ratings were collected. We removed a small number of responses that were obviously invalid (e.g., respondents giving all 1’s or all 10’s without variation, which we suspected was not genuine). 

\subsection{Results}

\begin{table}[h]
\centering
\small
\begin{tabular}{l|c|l|c}
\toprule[1.2pt]
Language & BLEU & Language & BLEU \\
\midrule[1.2pt]
yor & 0.4460 & kin & 0.5978 \\
amh & 0.6307 & lua & 0.3268 \\
afr & 0.3688 & kon & 0.3496 \\
ibo & 0.4945 & bem & 0.3571 \\
lin & 0.2163 & dik & 0.3714 \\
hau & 0.4434 & kik & 0.2432 \\
cjk & 0.4696 & ewe & 0.3887 \\
lug & 0.4468 & kam & 0.4896 \\
fuv & 0.4793 & kmb & 0.5262 \\
kab & 0.8565 & dyu & 0.5646 \\
\bottomrule[1.2pt]
\end{tabular}
\caption{BLEU scores for the AfriCaption dataset across 20 African languages. Language codes follow ISO 639-3 standards. 
\label{tab:bleu_scores}}
\end{table}

Table~\ref{tab:bleu_scores}  shows our effort to evaluate automatic translations using BLEU, which are commonly applied metrics in MTs \citep{Papineni2002BleuAM, popovic-2015-chrf}. Although these metrics capture broad quality trends, we observe considerable variance in translation quality across languages when evaluated using BLEU (some languages score relatively high while others remain lower). A key factor is that many MT systems perform better in the forward direction (Eng → target) than in reverse (target → Eng). Consequently, our back-translation approach may yield artificially low scores, especially for morphologically complex or extremely low-resource languages \citep{graham2019translationesemachinetranslationevaluation}. We therefore complement BLEU with semantic similarity checks and human evaluations for a more robust quality assessment.

\textbf{Analysis on Human Evaluation}
We found Yoruba to have the highest quality of translation based on human evaluation, Hausa being the worst, and Igbo and Ewe intermediate (Figure ~\ref{fig:average_human_score_per_lang}, left). The standard deviation of scores was about 2.5–3.0 for all, showing quite a spread of opinions or varying quality across different captions.

Yoruba demonstrated the highest consistency, with an average ICC of 0.68 (moderate agreement). Igbo and Hausa showed lower agreement, with ICCs of 0.52 and 0.41, respectively. Categorical agreement mirrored ICC trends. Yoruba achieved Fleiss' kappa $\kappa = 0.32$ (moderate), while Igbo and Hausa scored $\kappa = 0.32$ and $\kappa = 0.32$, respectively. 

The moderate agreement for Yoruba aligns with its relatively robust machine translation (MT) pipelines (NLLB and Azure Translate) and syntactic simplicity. For instance, the phrase “red-seated swing” translated smoothly as “ìyípadà ìjókòó pupa” (Yoruba), receiving 78\% excellent ratings. In contrast, Hausa’s low agreement correlates with grammatical errors (e.g., “kayaks” mistranslated as “teku,” a general term for “sea”) and limited MT training data (Costa-Jussa et al., 2022). Igbo’s bimodal scores likely stem from inconsistent handling of idiomatic phrases, such as “taking a swing” translated literally as “ewere swing” (Igbo), which 41\% of raters deemed Poor.

\begin{table*}[t]
\centering
\footnotesize
\begin{tabular}{l|cccccccccc}
\toprule[1.2pt]
 & \multicolumn{10}{c}{Languages} \\
\cmidrule(lr){2-11}
 & afr & amh & hau & ibo & lug & lin & kin & yor & cjk & dyu \\
\midrule[1.2pt]
BLEU   & 71.12 & 41.03 & 22.24 & 39.35 & 32.74 & 39.03 & 18.28 & 34.85 & 0.92 & 0.72 \\
ChrF++ & 82.32 & 61.96 & 42.20 & 60.85 & 54.54 & 59.68 & 38.21 & 55.56 & 15.03 & 13.31 \\
\midrule
\rowcolor{gray!15}
 & dik & ewe & fuv & kam & kab & kmb & kik & kon & lua & bem \\
\midrule
BLEU   & 4.22 & 1.08 & 1.31 & 1.55 & 0.50 & 0.44 & 1.68 & 0.16 & 0.76 & 1.16 \\
ChrF++ & 20.24 & 14.57 & 15.89 & 17.64 & 14.26 & 13.82 & 17.29 & 14.77 & 15.51 & 16.64 \\
\bottomrule[1.2pt]
\end{tabular}
\caption{Translation quality across languages measured by BLEU and ChrF++ scores. Language codes (e.g., amh for Amharic, afr for Afrikaans, etc.) follow ISO 639-3 standards. Full language definitions are provided in Appendix~\ref{tab: languages_and_definitions}.}
\label{tab: bleu_chrf}
\end{table*}
Yoruba translation misinterpreted “taking a swing”; a tricky idiom, leading to confusion. Similarly in Hausa, “Three people participate in rock climbing.” received a low ~4.4 average; the Hausa translation apparently lost the idea of “rock climbing” (perhaps translating literally in a strange way). On the other hand, Hausa raters gave ~8.6 on average to “Women walking down the street.”, indicating that simple captions were handled well. Igbo showed a polarized trend: several captions were rated very high ($\sim$8.0–8.3) but a few were low ($\sim$4.6–5.4). This suggests the Igbo MT sometimes produced excellent results and sometimes failed, perhaps due to inconsistent training data coverage for certain vocabulary. Ewe data is too sparse to draw strong conclusions, but interestingly, the two Ewe evaluators disagreed on many items (one gave much lower scores than the other), reflecting subjectivity or possibly differences in dialec. The human evaluation performed validates that \afri machine-translated captions are generally understandable and contextually relevant, though not flawless. They provide a realistic testbed: models trained on or evaluated against these captions will encounter some “noise” or errors as evidenced in our model's output.

\section{\afri Model}
\label{sec:model}

\begin{figure}[t]
  \centering
  \includegraphics[width=0.5\textwidth]{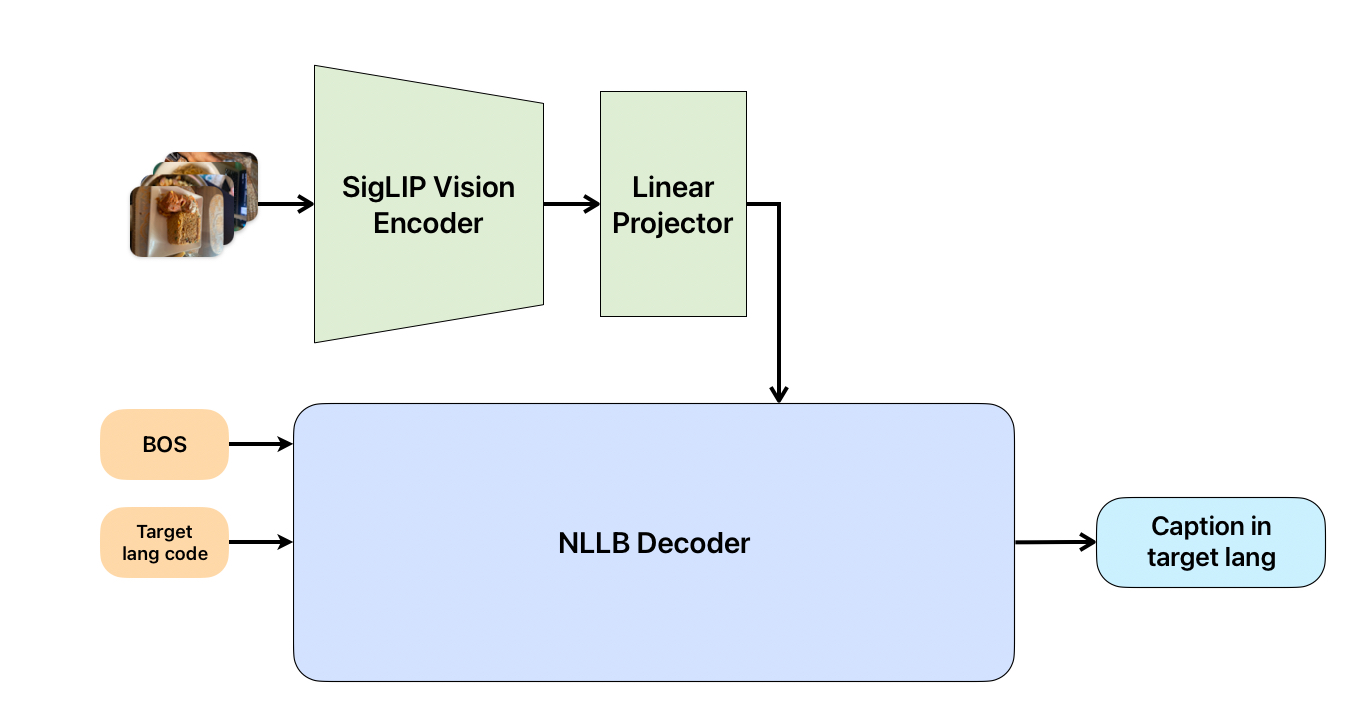}
  \caption{\afri model architecture.}
  \label{fig:model}
\end{figure}

\afri model is a vision-encoder–text-decoder model that integrates a pre-trained vision encoder with a pre-trained sequence-to-sequence language model’s decoder (Figure~\ref{fig:model}), designed specifically for multilingual image captioning in low-resource African languages. Given an input image and a designated language code, \afri generates captions autoregressively, producing text in the specified target language. Our model is capable of generating captions in up to 20 African languages listed in Section~\ref{sec:dataset}, thereby addressing a critical gap in image captioning for low-resource languages. \afri consists of three components: SigLIP's Vision Encoder, NLLB Decoder and a linear projector.

\subsection{Encoder}
% \quad
For our vision encoder, we use the publicly available multilingual variant of SigLIP's \citep{zhai2023sigmoid} image encoder, which is tailored to support multiple languages. This model employs sigmoid loss instead of softmax loss for contrastive pretraining of image-text pairs, demonstrating state-of-the-art performance, particularly given its small size

\subsection{Decoder}
% \quad
For our text decoder, we use the publicly available NLLB \citep{costa2022no} checkpoint, which covers 200 of the world's spoken languages (20 of which \afri focuses on). In our setup, the NLLB decoder generates a sequence of wordpiece tokens conditioned on the visual features extracted by the SigLIP encoder. The NLLB decoder produces output sequentially and employs an attention masking mechanism that restricts each generated token to attend only to previously generated tokens, thereby ensuring an autoregressive generation process. NLLB’s tokenizer handles language-specific tokenization. \\

\subsection{Vision-Encoder–Text-Decoder Integration}
\quad
The image encoder and text decoder are integrated using a modified version of Hugging Face’s VisionEncoderDecoderModel class. The visual features produced by the encoder are projected to match the decoder’s hidden size, ensuring compatibility when performing encoder–decoder cross-attention. During training, the model prepares the decoder’s input by shifting the target sequence to the right—ensuring that each output token only attends to preceding tokens, as required in sequence-to-sequence learning. Finally, an lm\_head linear layer is applied to project the decoder's hidden states to the size of the vocabulary, and a softmax function produces the probability distribution over the target tokens. This design allows for seamless encoder–decoder cross-attention and end-to-end training, and it is relevant for our task of choosing, image captioning.

\subsection{Training}
The training of \afri follows a two-stage fine-tuning training technique, which we detail in this section. 

\textbf{Stage 0: Selective layer pretraining}
Firstly, we take the publicly available checkpoints of the pretrained models off-the-shelf and integrate them using a custom Hugging Face VisionEncoderTextDecoder class to include an LM Head at the final layer of the NLLB decoder \citep{costa2022no}. We train the last layer of the vision encoder model along with the linear projection layer with the aim of aligning the image and text modalities. SigLIP \citep{zhai2023sigmoid} traditionally uses an encoder language model; however, most language models with African language translation capabilities are either decoder-only transformers or encoder-decoder language models. We opt for the NLLB decoder, as it has decent African language translation capabilities compared to other multilingual language models. We train for 40 epochs, using an lr of 2.0e-5 and a batch size of 16 on an L4 GPU. 

\textbf{Stage 1: Multimodal Pretraining}
In this stage, we pretrain the resulting model from Stage 1 for the image captioning task. The goal is to have a model that has acquired image captioning skills and be able to generate correct image captions in 20 African languages. We do not freeze any layer in our models like we did in the first stage. It is common practice to keep the image encoder frozen during this stage due to findings in LiT [132], reporting multimodal tuning of pretrained image encoders degrading their representations. Studies like CapPa \citep{tschannen2023image} and \citep{wan2024locca} have shown that captioning tasks can provide valuable signals to image encoders, allowing them to learn spatial and relational understanding capabilities that contrastive models like CLIP or SigLIP typically lack. Hence, we do not freeze the image encoder. We use a slow linear warm-up for our learning rate and an inverse root decay after the warm-up phase, which helps to first stabilize training (via warm-up) and then maintain a slowly decreasing learning rate to allow the model to train its parameters over time. We train for 30 epochs using an lr of 2.0e-5 and a batch size of 16 on an L4 GPU. \\

\subsection{Results and Analysis}
The model demonstrated steady improvement across training epochs, as evidenced by the progressive reduction in both training loss and validation loss, indicating an overall improvement in model confidence and generalization.

\begin{table}[h]
\centering
\scriptsize
\begin{tabular}{llrrlrrrrrrr}
\toprule
\textbf{Model} & \textbf{Lang} & \textbf{Bleu} &  \textbf{Cider} & \textbf{Spice} \\
\midrule

        Pangea & amh & 0 & 2.642e-08 & 2.750e-3 \\
               & igb & 0.0014 & 1.127e-07 & 4.981e-3 \\
        \midrule  
        AfriCaption & afr & 0.8387 & 8.3207 & 0.8358 \\
               & amh & 0.7768 & 7.6630 &  0.7906 \\
               & bem & 0.4813 & 4.4306 & 0.4952 \\
               & cjk & 0.2167 & 1.7945 & 0.1977 \\
               & dik & 0.2521 & 2.0666 & 0.2009 \\
               & dyu & 0.1732 & 1.3900 & 0.2288 \\
               & ewe & 0.2262 & 1.6527 & 0.1790 \\
               & fuv & 0.3552 & 2.1234 & 0.2192 \\
               & hau & 0.8567 & 8.4716 & 0.8435 \\
               & ibo & 0.8506 & 8.4087 & 0.8433 \\
               & kab & 0.1019 & 0.7688 & 0.0899 \\
               & kam & 0.1691 & 1.2574 & 0.1493 \\
               & kik & 0.1844 & 1.5548 & 0.1538 \\
               & kin & 0.7753 & 7.5940 & 0.7791 \\
               & kmb & 0.2407 & 1.9489 & 0.1968 \\
               & kon & 0.4129 & 3.4451 & 0.3728 \\
               & lin & 0.4044 & 3.6024 & 0.3807 \\
               & lua & 0.3017 & 2.6935 & 0.3162 \\
               & lug & 0.5156 & 5.0911 & 0.5364 \\
               & yor & 0.8212 & 7.9930 & 0.8127 \\
\bottomrule
\end{tabular}
\caption{Performance Comparison between Pangea and \afri per language.}
\label{table:detailed_table}
\end{table}

% \begin{table*}[t]
%   \centering
%   \small
%   \renewcommand{\arraystretch}{1.5} % Adjust row spacing
%   \setlength{\tabcolsep}{8pt}       % Adjust column spacing
%   \begin{tabular}{l m{5cm} m{5cm}} 
%     \toprule
%     \textbf{Lang} & \textbf{Reference} & \textbf{Prediction} \\
%     \midrule
%   yor &
% Ob\`{\i}nrin kan t\'{o} n\'{\i} \`{a}p\`{o} \d{\`{e}}y\`{\i}n r\d{\`{e}} j\'{o}k\`{o}\`{o} s\'{o}r\'{\i} \`{a}p\'{a}ta \'{n}l\'{a} kan, \'o s\`{\i} wo \'{a}w\textsubdot{o}n \`{o}k\`{e} \'{n}l\'{a}. (A woman with backpack sits on a large rock and looks down over the mountains.) 
% &
% Ob\`{\i}nrin kan t\'{o} n\'{\i} \`{a}p\`{o} \d{\`{e}}y\`{\i}n \`{j} j\'{o}k\`{o}\`{o} l\'{o}r\'{\i} \`{a}p\'{a}ta \'{n}l\'{a} kan t\'{o} \'o s\`{\i} wo \`{o}k\`{e} \`{o}k\`{e} \'{n}l\'{a} t\'{o} (A woman with \textcolor{yellow}{a} backpack sat on a large rock and looked \textcolor{yellow}{up at the mountain}.) \\
%     \midrule    
%     amh & {\amharicfont በህዝብ መካከል አንዲት ልጃገረድ የተንቆጠቆጠች በግ ይዛ}!(A girl in a crowd is holding on to a leashed sheep!) & {\amharicfont በህዝብ ውስጥ አንዲት ወጣትጃገረድ በንቆጠቆጠውንውን ለመዛ ተሸ}  (A young girl in the crowd is holding \textcolor{red}{a fringe}) \\
%     \midrule
%     hau & Mata biyun sun kwana a kan ciyawa tare (The two women slept on the grass together) &  Mata biyuun sun kwana a ci ciyawa (The two women slept on \textcolor{red}{[ci]} grass) \\
%     \bottomrule
%   \end{tabular}
  
%   \caption{Comparison of the groundtruth translations and our models output for 3 African Languages}
%   \label{tab:models output}
% \end{table*}

\begin{table*}[t]
  \centering
  \small
  \renewcommand{\arraystretch}{1.5} % Adjust row spacing
  \setlength{\tabcolsep}{8pt}       % Adjust column spacing

  {\centering
  \includegraphics[width=0.95\linewidth]{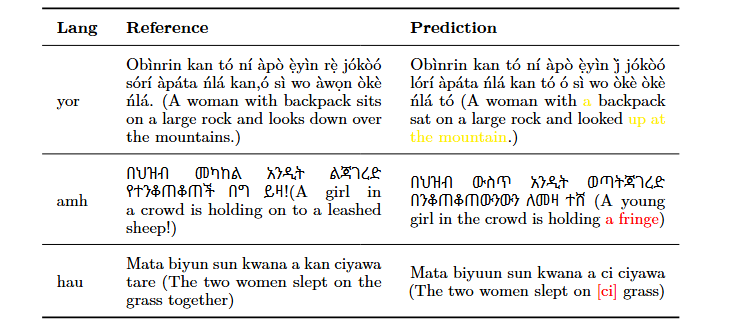}
  }

  \caption{Comparison of the groundtruth translations and our model's output for 3 African Languages}
  \label{tab:models_output_screenshot}
\end{table*}

Table~\ref{table:detailed_table} presents a performance comparison between the \afri model and Pangea model, a state-of-the-art, open-weight, multilingual multimodal model across BLEU, CIDEr, and SPICE metrics.
Table \ref{tab:models_output_screenshot} shows our models' output and it effectively captures the context of the images and generates complete sentences. In some instances, it produces words that are semantically similar to those in the ground truth captions, while in others, it omits one or two words within the caption. For the English translations, we highlight missed or unrelated words in red, indicating that they do not align with the image or the ground truth caption. Words that are contextually similar such as verb tense variations (e.g., a present-tense verb in the ground truth appearing in past tense in the model's output) are marked in yellow to reflect their near-equivalence in meaning.
 
\section{Discussion}

Our results show that the model is capable of generating image captions in a variety of African languages, achieving high-quality outputs in some cases while facing challenges in others. It consistently outperforms the Pangea model across the two overlapping languages; see the detailed table in Table \ref{table:detailed_table}. However, performance still varies, revealing broader limitations in existing tools for African language processing. These findings underscore the ongoing need for more robust multilingual AI systems, particularly for low-resource settings.

A key factor contributing to this disparity is the uneven representation of African languages in current “massively multilingual” MT models. Languages like Hausa and Yoruba, which have a relatively stronger digital presence and were likely better represented in training data, yielded better results compared to languages like Ewe or Dinka. This suggests that not all African languages benefit equally from such models, reinforcing the need for more inclusive and balanced training datasets. This raises an important question: \textit{can we bootstrap better translations by leveraging closely related languages?} For example, the strong performance in Luganda, a Bantu language, suggests the potential to improve captions for other Bantu languages like Zulu or Xhosa if extended to those languages. Our dataset provides a benchmark for such explorations, offering a foundation for testing fine-tuned MT models on captioning tasks across diverse African languages.

\section{Conclusion}
We introduced \afri, a family of the multilingual multimodal model that generates image captions in 20 African languages and the dataset that consists of 8k image-text samples in 20 African languages and English. Together, the model and dataset address the problem of exclusion of African languages in the vision-language domain, laying the foundation for broader inclusivity in multimodal AI. AfriCaption serves as a foundation for future research in multilingual image captioning and multimodal learning. We make our dataset and model available publicly on hugging face and we hope this spurs the development of more inclusive AI models that can understand and caption images in the languages spoken by the different communities in Africa.

Moving forward, we plan to adopt a participatory approach, similar to Masakhane, to refine and validate captions. We also advocate incorporating culturally specific imagery and descriptions to ensure models resonate with diverse African contexts. Ultimately, AFRICaption is a pivotal step toward bridging the multimodal resource gap and fostering equitable, multilingual AI systems.

\section{Limitations}

While \afri significantly advances multilingual AI inclusivity, it also highlights systemic gaps in low-resource language research. Translation quality remains uneven, for example, Yoruba outperforms languages like Hausa, Ewe, or Dinka due to richer digital representation, and standard back-translation evaluation metrics (e.g., BLEU) often miss semantic nuances in morphologically complex languages. Furthermore, limited community involvement in human evaluation may overlook subtle, culturally nuanced errors.

Beyond these methodological gaps, this work also lacks cultural awareness. While our dataset and models represent a step toward enabling image captioning in African languages for basic daily conversations, they do not yet capture the deeper cultural context embedded in language use, such as idiomatic expressions, social norms, or culturally specific references. Future iterations of this work would benefit from stronger integration of cultural perspectives, ensuring that captions reflect not only linguistic accuracy but also the lived realities of African communities.

\bibliography{acl}

\appendix
\section{Appendix}
\label{sec:appendix}

% \subsection{
\textbf{Survey on Dataset - Human Evaluation}
\label{sec: survey}

\begin{figure}[!htbp]
  \centering
  \includegraphics[width=0.8\linewidth]{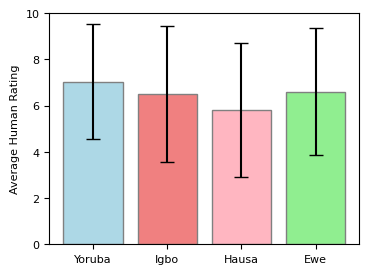}
  \caption{
    A plot of the \textbf{average human score per language} with error bars denoting standard deviation.
  }
  \label{fig:average_human_score_per_lang}
\end{figure}

We created evaluation surveys where native speakers were presented with the original English caption and the translated caption in their language. For each caption pair, we asked evaluators to rate the translation’s adequacy on a scale from 1 to 10, with instructions that 1 means “completely wrong translation,” 5 means “understandable gist but with errors,” and 10 means “perfect translation that preserves the full meaning.” We also asked them to flag any catastrophic errors, like when the translation says something entirely different from the original caption.

\begin{figure*}
 \centering
 \includegraphics[width=\textwidth]{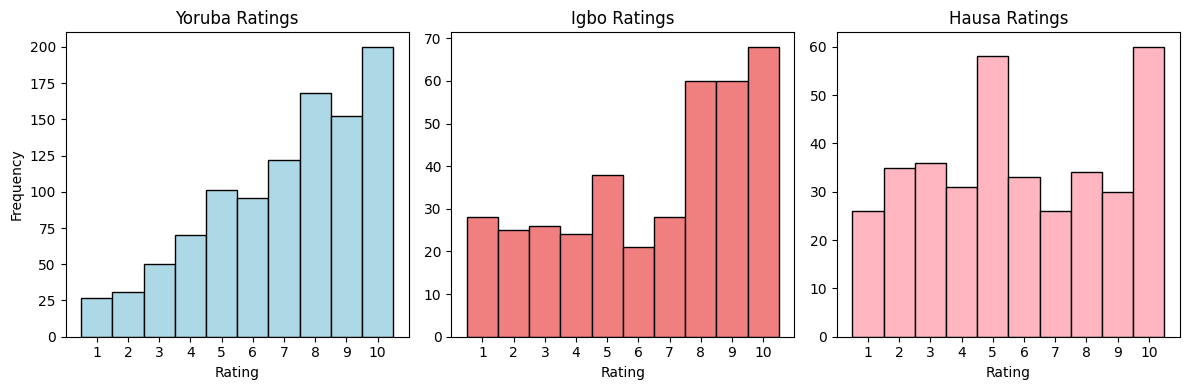}
 \caption{Score distributions for Yoruba, Igbo, and Hausa. We observe that over 50\% of Yoruba ratings were 8 or above, and $\sim $20\% were perfect 10s. Hausa’s distribution is flatter, with a mode around 10 (16\% of scores were 10) but also a substantial portion of low scores (1–4 ratings made up $\sim $35\% of Hausa responses, compared to only  $\sim $18\% for Yoruba). Igbo’s distribution is bimodal – it has a high incidence of 9–10 scores (about one-third of Igbo ratings were 9 or 10, similar to Yoruba) and a noticeable chunk of very low scores (1’s, 2’s, 3’s accounted for $\sim $21\% in Igbo, versus $\sim $9\% in Yoruba). This bimodality aligns with the earlier observation of Igbo translations being hit-or-miss}
 \label{fig: human_eval_histogram}
\end{figure*}

\begin{figure}[H]
  \centering
  \includegraphics[width=0.8\linewidth]{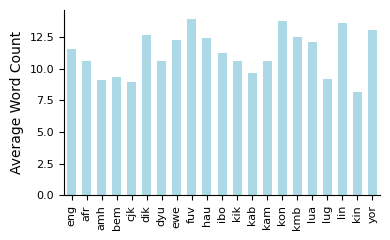}
  \caption{
    Average word count per language in the AFRICaption dataset. 
    The plot highlights variations in caption lengths across different languages.}
  \label{fig:average_word_count}
\end{figure}

\textbf{Perceived Data Quality vs. Average Length}
\label{sec: perceived_quality_on_length}
Figure~\ref{fig:average_word_count} compares the average word count per language caption in our dataset. It shows that our dataset achieves a reasonable balance in caption length across 20 African languages, with the English captions providing a baseline.
The consistent average word counts suggest that the translations are neither too brief nor overly verbose, preserving the essential information while ensuring readability. According to previous studies \citep{singh2024aya}, balanced caption length is a key feature in preventing model bias in and improving interpretability. This characteristic makes our dataset well-suited for training models that need to generalize across diverse linguistic contexts.

% Second figure

\begin{table*}
    \centering
    \small
    \begin{tabular}{ @{} c  l  l @{} }
        \hline
        \textbf{ISO 639-3} & \textbf{Language Name} & \textbf{Countries (with Flags)} \\ 
        \hline \\
        afr & Afrikaans & \smallflag{ZA} South Africa, \smallflag{NA} Namibia \\[4pt]
        amh & Amharic & \smallflag{ET} Ethiopia \\[4pt]
        hau & Hausa & \smallflag{NG} Nigeria, \smallflag{NE} Niger, \smallflag{GH} Ghana, \smallflag{TD} Chad, \smallflag{CM} Cameroon \\[4pt]
        ibo & Igbo & \smallflag{NG} Nigeria \\[4pt]
        lug & Luganda & \smallflag{UG} Uganda \\[4pt]
        lin & Lingala & \smallflag{CD} DR Congo, \smallflag{CG} Rep. Congo, \smallflag{AO} Angola, \smallflag{CF} Central African Rep. \\[4pt]
        kin & Kinyarwanda & \smallflag{RW} Rwanda, \smallflag{CD} DR Congo, \smallflag{UG} Uganda \\[4pt]
        yor & Yoruba & \smallflag{NG} Nigeria, \smallflag{BJ} Benin, \smallflag{TG} Togo \\[4pt]
        cjk & Chokwe & \smallflag{AO} Angola, \smallflag{CD} DR Congo \\[4pt]
        dyu & Dyula (Jula) & \smallflag{BF} Burkina Faso, \smallflag{CI} Côte d'Ivoire, \smallflag{ML} Mali \\[4pt]
        dik & Dinka & \smallflag{SS} South Sudan \\[4pt]
        ewe & Ewe & \smallflag{GH} Ghana, \smallflag{TG} Togo \\[4pt]
        fuv & Fulfulde (Fula) & \smallflag{NG} Nigeria, \smallflag{CM} Cameroon, \smallflag{GN} Guinea, \smallflag{SN} Senegal, \smallflag{ML} Mali \\[4pt]
        kam & Kamba & \smallflag{KE} Kenya \\[4pt]
        kab & Kabyle & \smallflag{DZ} Algeria \\[4pt]
        kmb & Kimbundu & \smallflag{AO} Angola \\[4pt]
        kik & Kikuyu & \smallflag{KE} Kenya \\[4pt]
        kon & Kongo & \smallflag{CD} DR Congo, \smallflag{CG} Rep. Congo, \smallflag{AO} Angola \\[4pt]
        lua & Luba-Kasai & \smallflag{CD} DR Congo \\[4pt]
        bem & Bemba & \smallflag{ZM} Zambia \\[4pt]
        \hline
    \end{tabular}
    \caption{Languages and their definitions.}
    \label{tab: languages_and_definitions}
\end{table*}

\begin{table*}
\centering
\resizebox{1.0\textwidth}{!}{
\begin{tabular}{l|cccccccccc}
\toprule[1.2pt]
 & \multicolumn{10}{c}{Languages} \\
\cmidrule(lr){2-11}
 & afr & amh & bem & cjk & dik & dyu & ewe & fuv & hau & ibo \\
\midrule[1.2pt]
No. of Tokens & 135336 & 148517 & 58640 & 19805 & 16042 & 2021 & 4734 & 8126 & 127429 & 152722 \\
\rowcolor{gray!15}
 & kik & kab & kam & kon & kmb & lua & lug & lin & kin & yor \\
No. of Tokens & 12494 & 2250 & 6035 & 15467 & 2757 & 16974 & 66018 & 22109 & 120542 & 172016 \\
\midrule
No. of Characters & 425977 & 294622 & 206608 & 75104 & 51995 & 6745 & 14967 & 27041 & 478142 & 478600 \\
\rowcolor{gray!15}
 & kik & kab & kam & kon & kmb & lua & lug & lin & kin & yor \\
No. of Characters & 41032 & 6417 & 19750 & 60924 & 9890 & 67607 & 219135 & 88618 & 420213 & 461489 \\
\midrule
Avg. Length & 3.15 & 1.98 & 3.52 & 3.79 & 3.24 & 3.34 & 3.16 & 3.33 & 3.75 & 3.13 \\
\rowcolor{gray!15}
 & kik & kab & kam & kon & kmb & lua & lug & lin & kin & yor \\
Avg. Length & 3.28 & 2.85 & 3.27 & 3.94 & 3.59 & 3.98 & 3.32 & 4.01 & 3.49 & 2.68 \\
\bottomrule[1.2pt]
\end{tabular}}
\caption{Statistical overview of language characteristics. Language codes: afr (Afrikaans), amh (Amharic), bem (Bemba), cjk (Chokwe), dik (Dinka), dyu (Dyula), ewe (Ewe), fuv (Fulfulde), hau (Hausa), ibo (Igbo), kik (Kikuyu), kab (Kabyle), kam (Kamba), kon (Kongo), kmb (Kimbundu), lua (Luba-Katanga), lug (Luganda), lin (Lingala), kin (Kinyarwanda), yor (Yoruba).}
\label{tab:language_stats_final}
\end{table*}

\end{document}